\documentclass[conference]{IEEEtran}
\IEEEoverridecommandlockouts
\usepackage{cite}
\usepackage{amsmath,amssymb,amsfonts}
\usepackage{algorithmic}
\usepackage{graphicx}
\usepackage{textcomp}
\usepackage{xcolor}
\usepackage{epstopdf}
\usepackage{CJK}

\def\BibTeX{{\rm B\kern-.05em{\sc i\kern-.025em b}\kern-.08em
    T\kern-.1667em\lower.7ex\hbox{E}\kern-.125emX}}
\begin{document}

\title{CBOWRA: A Representation Learning Approach for Medication Anomaly Detection}

\author{\IEEEauthorblockN{Liang Zhao$^1$, Zhiyuan Ma$^{1,*}$, Yangming Zhou$^{1}$, Kai Wang$^{1}$, Shengping Liu${^2}$, Ju Gao$^3$ and Wen Du${^4}$}
\IEEEauthorblockA{$^1$School of Information Science and Engineering, East China University of Science and Technology, Shanghai 200237, China \\
$^2$ Unisound, Shanghai 200030, China \\
$^3$Shuguang Hospital Affiliated to Shanghai University of Traditional Chinese Medicine, Shanghai 200021, China \\
$^4$DS Information Technology Co.,Ltd, Shanghai 200032, China\\
$^*$Corresponding author\\
Emails: yuliar3514@gmail.com\\
}
}

\maketitle
%
%
\begin{CJK}{UTF8}{gbsn}
\begin{abstract}
Electronic health record is an important source for clinical researches and applications, and errors inevitably occur in the data, which lead to severe damages to both patients and hospital services. 
One of such errors is the mismatch between diagnose and prescription, which we address as ``medication anomaly'' in the paper, and clinicians used to manually identify and correct them. 
With the development of machine learning techniques, researchers are able to train specific model for the task, but the process still requires expert knowledge to construct proper features, and few semantic relations are considered. 
In this paper, we propose a simple, yet effective detection method that tackles the problem by detecting the semantic inconsistency between diagnoses and prescriptions. 
Unlike traditional outlier or anomaly detection, the scheme uses continuous bag of words to construct the semantic connection between specific central words and their surrounding context. 
The detection of medication anomaly is transformed into identifying the least possible central word based on given context. To help distinguish the anomaly from normal context, we also incorporate a ranking accumulation strategy. 
The experiments were conducted on two real hospital electronic medical records, and the $top N$ accuracy of the proposed method increased by 3.91 to 10.91\% and 0.68 to 2.13\% on the datasets, respectively, which is highly competitive to other traditional machine learning-based approaches.
\end{abstract}


\section{Introduction}
Electronic Health Record (EHR), is first motivated to curb the trend of increasing cost in health and medical care \cite{yadav2018mining,liu2018patienteg}. 
It contains patients' information like physical conditions, diagnoses, prescriptions, and is considered to have the greatest potential in elevating medical services.

Despite the fact that EHRs save considerable amount of financial cost, to use such records will have to deal with errors like unintended or wrong prescriptions \cite{zwaan2015challenges}, which sometimes cause severe damages for not only patients themselves, but also to hospital services. 
Table \ref{Tab:exa_error} is an example from a patient's EHR file. It can be noted that some of the medications is irrelevant to the diagnoses. For example, ``艾司唑仑片 (estazolam tablets)'' is used for anti-anxiety and insomnia, and ``氯硝西泮片 (clonazepam tablets)'' is for epileptic seizures. Clearly, the prescriptions do not match the diagnoses. 
\begin{table}[htbp]
\caption{An Example of Electronic Medication Record}
\begin{center}
\newcommand{\tabincell}[2]{\begin{tabular}{@{}#1@{}}#2\end{tabular}}
\begin{tabular}{c|c}
\hline
\textbf{Name}& \textbf{Content} \\
\hline
\textbf{\tabincell{c}{Patient's ID}} & E20140702E1 \\
\hline
\textbf{Diagnoses} & \tabincell{c}{胃癌 (gastric cancer), 气血亏虚 (qi-blood \\deficiency),  胃底癌术后 \\(syndrome and fundus cancer after operation)} 	\\
\hline
\textbf{Prescriptions} & 
\tabincell{c}{
华蟾素注射液 (cinobufacini injection), \\参麦注射液 (shenmai injection) (50ml), \\ 养血饮口服液 (yangxueyin oral liquid),\\ 百令胶囊 (bailing capsules), \\ 百士欣胶囊 (Baishixin capsules), \\ 多酶片 (multienzyme tablets), \\ 艾司唑仑片 (estazolam tablets), \\ 氯硝西泮片 (clonazepam tablets), \\ 培菲康胶囊 (bifikang capsules), \\香连片(fragrant pills), \\ 宁泌泰胶囊 (ningbitai capsules), \\ 金水宝胶囊 (jinshuibao capsules), \\ 消痛贴膏 (analgesic plaster), \\螺旋藻片 (spirulina tablets),\\ 开塞露 (kaiseru).}\\
\hline
\end{tabular}
\label{Tab:exa_error}
\end{center}
\end{table}

This type of error was first discussed in \cite{C1997Medication} by the term ``medication error'', and was categorized with wrong labeling, dispensing, dose, administration and preparation of drugs. 
Since inappropriate actions over drugs may cause severe damage to patients, researches related to such errors also belong to ``drug safety''. Moreover, the causes of the errors also owe to inappropriate diagnoses, which leads to part of the errors falling into the category of ``diagnostic error'' \cite{schiff2009diagnostic}.

Detecting such errors in EHRs is conventionally formulated as an outlier detection problem. Current researches of outlier detection can be generally split into distance-based and density-based methods \cite{aggarwal2015outlier}. The former is mainly based on Nearest Neighbor (NN) to implement a distance-metric for detection \cite{tran2016distance}, and the latter relies on Local Outlier Factor (LOF) to assign different degrees for each data sample \cite{li2015privacy}. 
In recent years, researchers have been studying and evaluating different methods regarding to outlier detection, which gives industrial practitioners valuable insights of choosing different models \cite{swersky2016evaluation,campos2016evaluation}. Nevertheless, these studies focus more on the effectiveness and efficiency of the model than on specific tasks, and the performances rely on the scale as well as the distribution of the data \cite{domingues2018comparative}.
In the meantime, learning the difference between outliers and normal samples cannot be done directly from the attributes, and the semantic meanings of the context should be considered. Taken medical fraud for example \cite{Capelleveen2016Outlier}, some of the frauds are coding or charging mistakes which should be excluded, while others like excessive or irrelevant services are to be detected. 

Traditional detection methods in medical domain rely on experts' effort to manually identify the errors, which is expensive and of low efficiency. Meanwhile, most of the researches mainly focus on practice guide \cite{goedecke2016medication}. 
With the developments in machine learning modeling, current approaches have been trying to achieve simple and expert-free detection. 
Schiff et al. \cite{schiff2009diagnostic} improved clinical decision system with machine learning-based approach which generates useful alert for medication error.
Later in 2017, they used outlier detection screening to help generate alerts for medication errors \cite{schiff2017screening}. However, the performance relies on MedAware system to provide the alert, and the medication errors are evaluated through chart review analysis, leading only three out of four alerts valid. 
Solving the problem by detecting the mismatches of medical records in feature space, Zhang et al. \cite{zhang2016probabilistic} proposed an anomaly detection paradigm based on extended Latent Dirichlet Allocation (LDA). 
But the performance relied on the quality of high dimensional features, and no contextual connection was considered. 
Other approaches have been targeting on specific disease \cite{taylor2018predicting}, and have not considered the semantic anomaly that resides in the context of the prescriptions. 
Taken Table \ref{Tab:exa_error} for example, ``开塞露 (kaiseru)'' is used for constipation, which mismatches with the patients' diagnoses. However, patients who have chemotherapy usually suffer from constipation as a side-effect, and clinicians always add Kaiseru to ease the pain. Hence, Kaiseru is not a medication error. The situation is similar to that in medication fraud detection, in which we cannot simply treat mismatches as anomalies. 
In the same time, mining the context of drug corpus has drawn attentions in recent years \cite{goh2018drug}, which motivates us to detect the errors through mining the semantic relationship between diagnoses and prescriptions.






As discussed in previous content, possible causes of the errors include wrong typing of prescriptions, misuse of the system, wrong diagnosis, etc. 
In the following part of the manuscript, we address the issue as medication anomaly. 
Clinical experts identify them based on domain knowledge. It means there exit certain semantic links among patients, diseases and drugs. Detecting the anomaly requires the machine to understand the meaning of the text, but current approaches regarding to natural language processing are mostly used for correcting syntactic errors \cite{cai2016natural,pons2016natural}. 
There is clearly a missing part of adopting NLP techniques to tackle the problem from the perspective of semantic connections between patients and drugs. 
To fill the gap, we were motivated to have the model learn the semantic inconsistency between the diagnoses and the prescriptions, and use the model to distinguish possible medication anomalies. 

Following the intuition, we proposed a representation learning method called Continuous Bag Of Words (CBOW) based on Ranking Accumulation (CBOWRA) to detect any possible anomaly in prescriptions. The method uses an Continuous Bag Of Words (CBOW) model to learn the representation of context, and an accumulated ranking strategy is included to help detect possible medication anomaly. 
The contributions of the manuscript are three folds.
\begin{enumerate}
\item We proposed a simple, yet effective representation learning method to detect medication anomaly based on patients' diagnoses and prescriptions. The training process of the model is expert-free, which saves considerable resources for practical adoption.
\item We modified conventional CBOW model in both structure and training procedure to predict the central word based on diagnoses. We also incorporate a ranking mechanism to help exaggerate the differences between normal drugs and the anomaly.
\item To illustrate the effectiveness of the proposed method, we designed detection process for non-representation and representation learning methods. The experimental results confirmed the superiority of the proposed method, which further provides insights for adopting representation learning models in similar scenarios.
\end{enumerate}

The rest of the paper is organized as follows. The detection scheme, along with the construction of the detection model, are detailed in Section 2. The evaluations of the proposed method are conducted on real EHR data in Section 3, in which the details of data preparing and comparative methods are included as well. Section 4 presents the discussions regarding to the proposed method and the conclusions are drawn as Section 5.

\section{Methodology}
\begin{figure}[htbp]
\centerline{\includegraphics[width=0.5\textwidth]{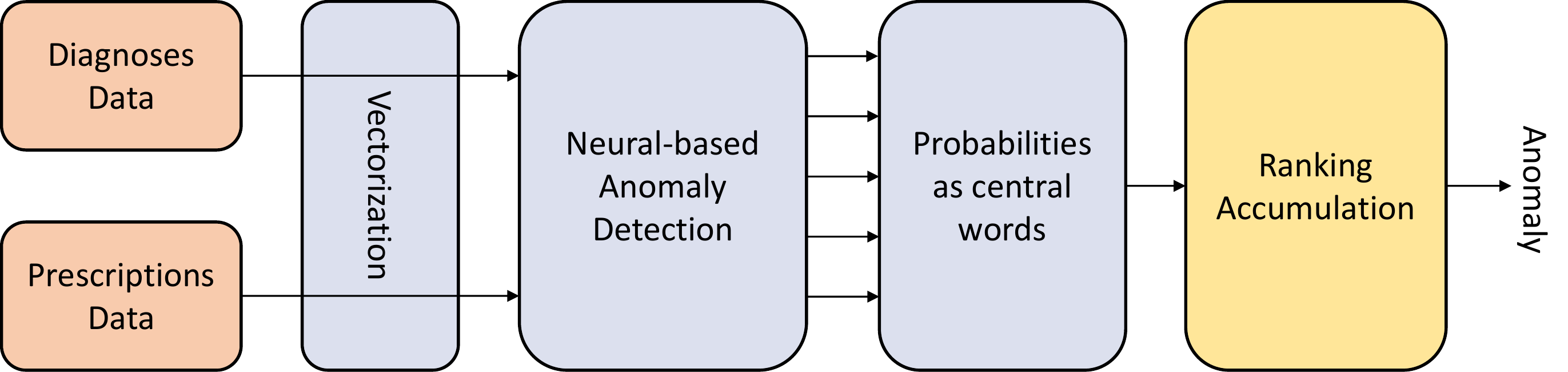}}
\caption{The outline of medication anomaly detection.}
\label{fig:Model_outline}
\end{figure}

The outline of the proposed method is given in Fig. \ref{fig:Model_outline}. 
Firstly, the diagnoses and prescriptions of each patient are vectorized and passed down through a neural-based anomaly detection model, i.e. CBOWRA, in which the semantic inconsistency between diagnoses and prescriptions will be evaluated. 
The output of the model is a $n$-length vector, where $n$ is the number of medications to be evaluated and each entry represents the possibility of corresponding medication as the central word. 
Multiple such groups of data are further passed to a ranking accumulation process, in which the entries are sorted in an ascending order. 
In this manner, the medication corresponding to the first entry is treated as an anomaly. 

To better illustrate the details of the proposed method, we split the following section into four parts. 
In subsection A
, we explain the vectorization of the diagnoses and prescriptions, followed by subsection B
, which provides a detailed description on the model. 
Considering the fact that the data are not in typical document format, we further describe the process of constructing the training set in subsection C
. In the end, the anomaly detection based on the proposed method is provided.

\subsection{Vectorization of Data}
\label{subsec:1}
\begin{figure}[htbp]
\centerline{\includegraphics[width=0.5\textwidth]{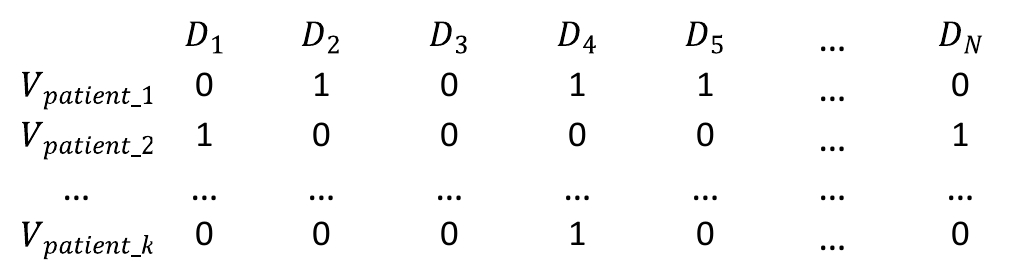}}
\caption{Examples of vectors for patients' diagnoses.}
\label{fig:multi_vec_patient}
\end{figure}

We consider two different codings corresponding to patients' diagnoses and prescriptions, respectively. 
For the former, we uses multi-hot vector, which is commonly used in outlier detection methods,  to represent the data. 
Specifically, each diagnosis is coded into an $N$-length vector, denoted as $V_{patient\_i},~(i=1,2,\cdots,k)$, where $k$ is the number of patients. Each entry represents the counts of corresponding disease appearing in the diagnosis. 
Fig. \ref{fig:multi_vec_patient} shows an example of such vectors. 

\begin{figure}[htbp]
\centerline{\includegraphics[width=0.5\textwidth]{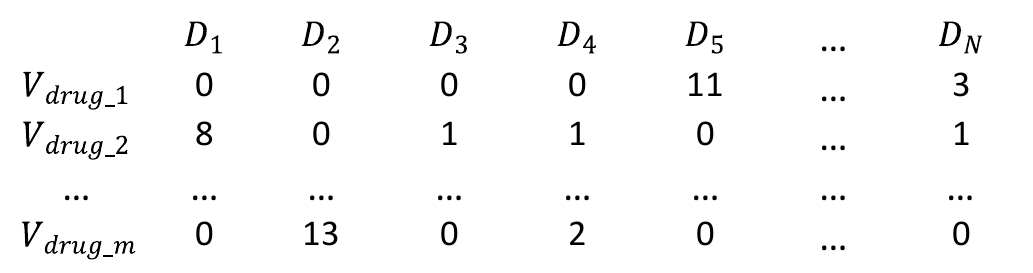}}
\caption{Examples of vectors for patients' prescriptions.}
\label{fig:bow_medication}
\end{figure}

As for the prescriptions, we organize raw data into a different form. 
Note that CBOW vectorizes the words based on vocabulary from the corpus. 
In terms of patients' prescriptions, each entry denotes the count number of the patients that have the medication in their prescriptions. Fig. \ref{fig:bow_medication} gives an example of such vectors for $m$ prescriptions. Note that, the number of patients that have disease $D_1$ and take $drug\_2$ is $8$, so the entry for $D_1$ of $V_{Drug\_2}$ is $8$. 

\subsection{Representation Learning based on CBOW}
\label{subsec:2}

Representation learning have been proved to be effective in revealing semantic information such as textual similarity\cite{yang2018learning}, semantic consistency \cite{gao2017video}, representations \cite{xie2018learning} and other context related features \cite{qiu2018fast}. Hence, we choose such learning model for mining the semantic inconsistency in patients files.


By evaluating current representation learning methods, including CBOW \cite{mikolov2013efficient}, TransE \cite{bordes2013translating}, TransH \cite{wang2014knowledge}, TransD \cite{ji2015knowledge}, and TransR \cite{lin2015learning}, 
we find that CBOW-based model obtained the best performance of the five. Therefore, we construct our representation learning model based on CBOW. 

\begin{figure}[htbp]
\centerline{\includegraphics[width=0.3\textwidth]{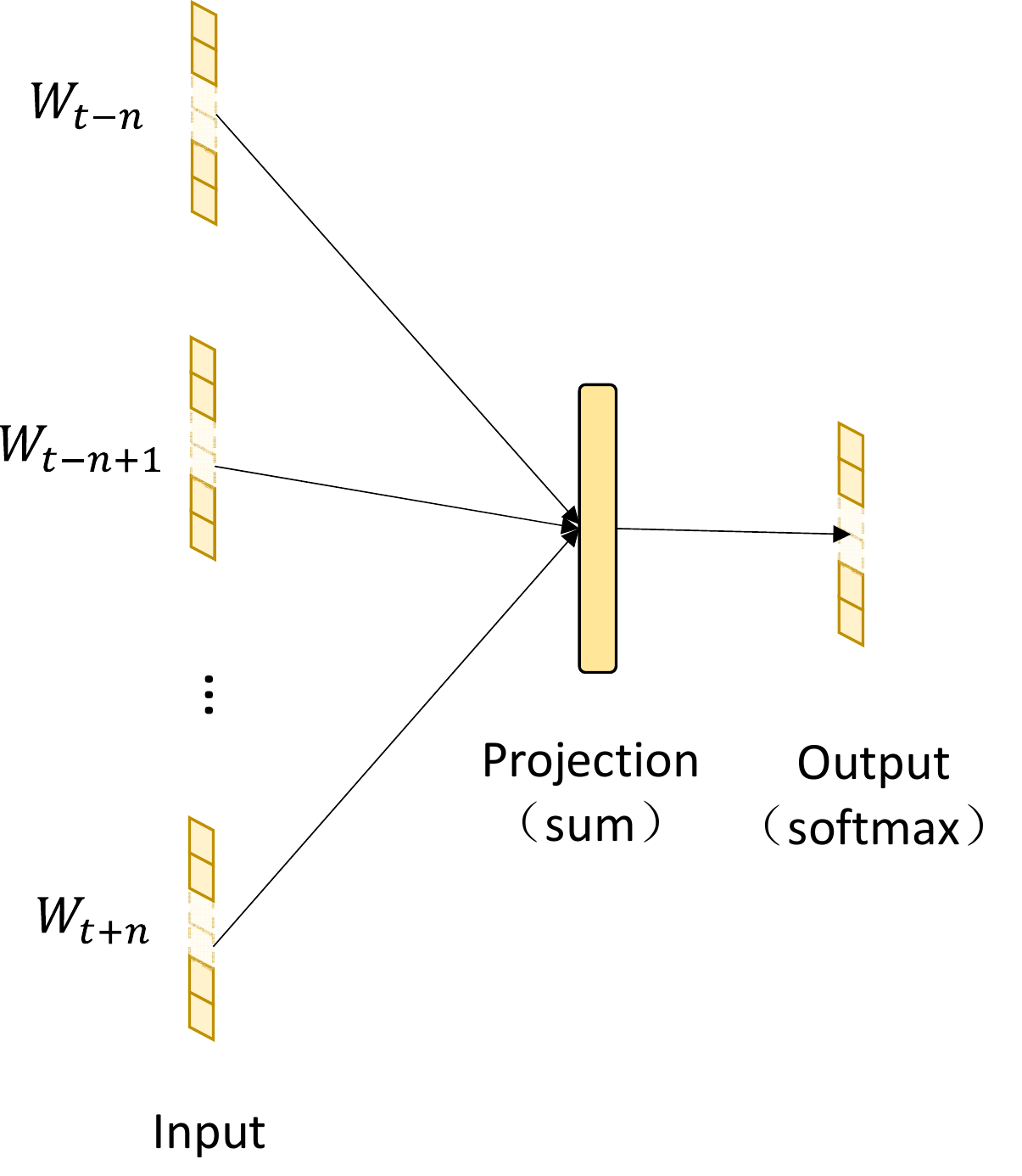}}
\caption{The network structure of CBOW.}
\label{fig:CBOW_conven}
\end{figure}

The evaluation results can be found in Section \ref{sec:exp}. 
The structure of CBOW can be shown as Fig. \ref{fig:CBOW_conven}. It is a feed forward neural network language model, in which non linear hidden layer is removed, leaving only a projection layer and a softmax layer \cite{mikolov2013efficient}.

The inputs $W_{t-n}, W_{t-n+1},\cdots,W_{t+n}$ denote the surrounding words that are presented in vector forms. The projection layer does a vector-wise summation of all inputs and passes the result to a softmax layer to produce the probability of each candidate word being the central word. 
The training of the model can be formulated as maximizing the probability of generating a correct central word, given by Equation (\ref{equ:3-1-1}). 
\begin{equation}
\label{equ:3-1-1}
\prod_{t=1}^{T} p(w^t | w^{(t-m)},\cdots,w^{(t-1)},w^{(t+1)},\cdots,w^{(t+m)})
\end{equation}

In the equation, $ w^{(t-m)},\cdots,w^{(t-1)},w^{(t+1)},\cdots,w^{(t+m)}$ are the surrounding words (or context), and $w^t$ is the central word to be predicted. The letter $T$ denotes the size of the dictionary, i.e. the number of words appeared in the corpus. $2m$ is the size of the sliding window used for training. 
The calculation of the probability $p$ is given as Equation (\ref{equ:3-1-2}), in which 
$v=\{v_{oi},~(i=1,2,\cdots)\}$ represents the vector set of the surrounding words and $u_c$ and $u_i$ are the vectors of the central words.
\begin{equation}
\label{equ:3-1-2}
\begin{aligned}
&p(w^t | w^{(t-m)},\cdots,w^{(t-1)},w^{(t+1)},\cdots,w^{(t+m)}) \\
& = \frac{\exp \left[ u_c^T (v_{o1} + \cdots + w_{o2m}) \right] /(2m)}{\sum_{i \in v} \exp \left[ u_{i}^T (v_{o1} + \cdots + w_{o2m})/(2m) \right] }
\end{aligned}
\end{equation}

Conventional CBOW has advantages such as light training burden, high quality word representation and so on. However, adopting CBOW to specific scenarios requires certain modifications. 
In the issue of medication anomaly detection, we are to consider two different types of documents, namely the patients-disease and disease-drug. 

Intuitively, medication anomaly happens at a relatively lower frequency when compared with appropriate or correct prescriptions. Therefore, it is reasonable to assume that the probability of wrong drug appearing in the diagnoses is lower than the normal or correct drugs. Since CBOW can predict the central word based on the context, we can use CBOW to generate corresponding probabilities for each drug, that is given to a specific patient, as the central word. Based on the aforementioned assumption, the medication with the lowest probability as the central word is most likely to be a prescription anomaly. Hence, the goal of detecting the anomaly becomes the detection of non-central words. 

\begin{figure}[htbp]
\centerline{\includegraphics[width=0.45\textwidth]{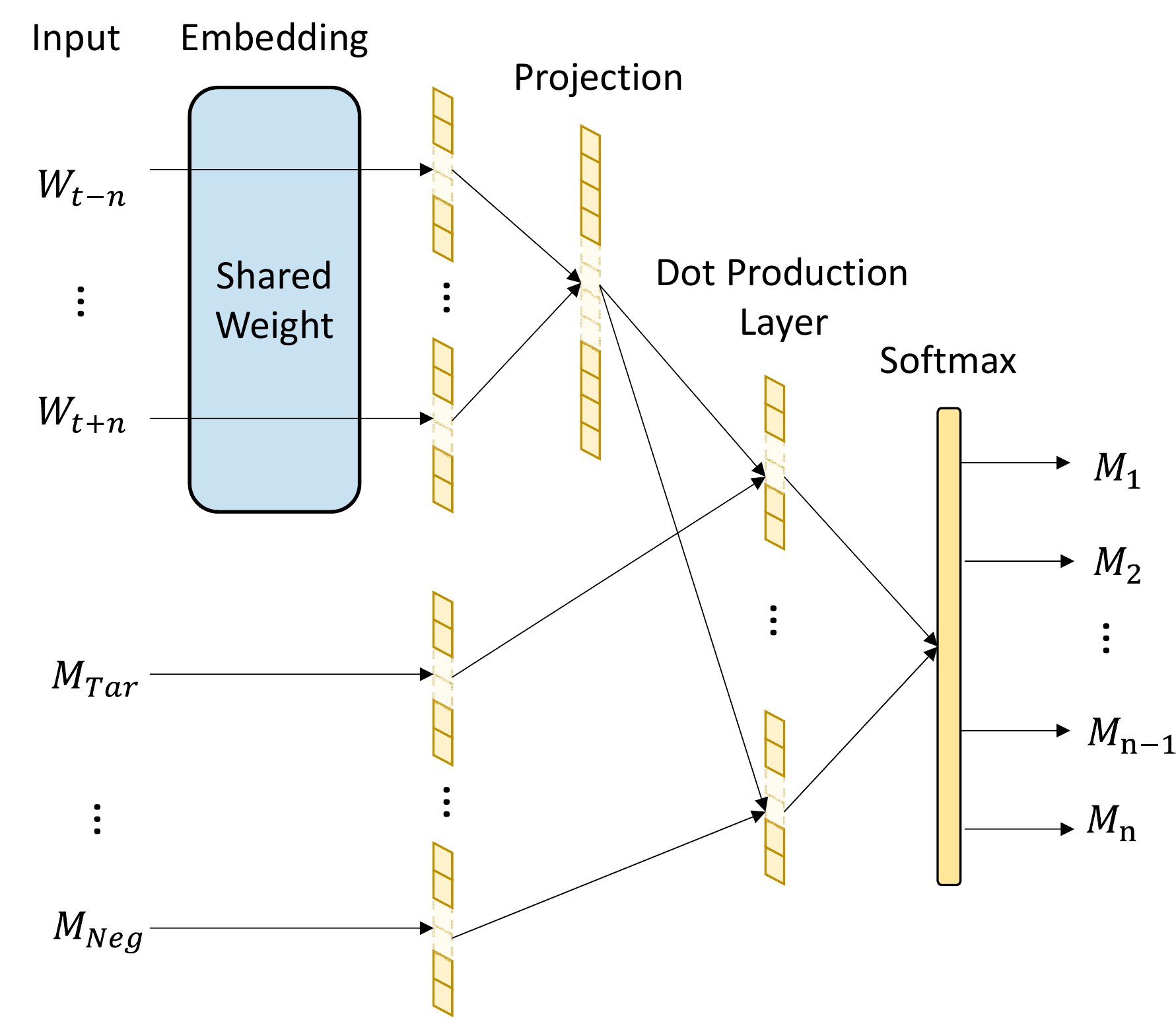}}
\caption{Network structure of CBOWRA.}
\label{fig:CBOW_PR}
\end{figure}

In this paper, we modified conventional CBOW into a multi-layer neural network, whose structure is given in Fig. \ref{fig:CBOW_PR}. 
The first layer is the input layer, where $W_{t-n},\cdots,W_{t+n}$ denote the surrounding words and $M_{Tar},\cdots,M_{neg}$ are the prescriptions given to the patient. 
Only surrounding words are transformed into word embeddings, before passing through a projection layer. The projection layer works in the same way as conventional CBOW does, which is to do a vector-wise summation. The proposed structure differs from conventional CBOW in the third layer, which performs a dot product of the output of the projection layer and the vectors of diseases. 
In this way, the coded information of the surrounding words in a specific patient's diagnoses will be emphasized by the prescription of the patient. Since the diagnoses and the prescriptions are vectorized according to the disease, it will help to identify the semantic inconsistency between a patient's diagnoses and prescription. 
In the end, the output of the third layer is passed to a softmax layer, and the probabilities of certain medications as central words are given by each entry of the output layer.

%

\subsection{Construction of Training Set}
\label{subsec:3}

To support the model described in the previous subsection, preparing the surrounding and central words is the next problem to solve. It is also well acknowledged that the disease shares similar or the same symptom(s) across diverse patients. Therefore, the construction or training of CBOW model requires to combine different patients' diagnoses. 

Similar to conventional CBOW, we use a sliding window, the size of which is denoted as $2m$ in Equation (\ref{equ:3-1-1}) and (\ref{equ:3-1-2}). 
In this way, each training sample contains $2m$ surrounding words and $1$ central word, which are extracted from the diagnoses of the patients.

For conventional CBOW, the middle word, denoted as $w_{m}$, is the central word. However, this is not the case for EHR data since the diagnoses and prescriptions are organized in a non-sequential manner. Let the number of medications in the prescriptions of one patient be $n$, and the number of diagnoses of the patient be $k$. To cope with the proposed method, we use the following steps to construct the training set. 
\begin{enumerate}
\item \label{step_1} From all the diagnoses of a patient, 
select $m$ different diagnoses to combine with the medication. In total, there will be $c_k^m$ surrounding words;
\item \label{step_2} Choose one drug from the prescription given to one patient as the central word, and combine it with the surrounding words generated in step $\left.\ref{step_1}\right)$.
\end{enumerate}

Note that conventional training process of a neural network involves updating all the weights, which results in heavy computing overhead. For example, let the dimension of the word vector be $300$, and the number of words be $20,000$. In this case, there will be an updating process for a $300 \times 20,000$ matrix. Performing gradient-based training can be very slow and additional amount of data are required to avoid over-fitting. 
To lighten the burden and simplify the training, we further incorporate negative sampling \cite{mikolov2013distributed} in the process.

The objective, given as Equation (\ref{equ:3-1-3}), is almost the same as \cite{mikolov2013distributed} except the meaning of terms are different.
\begin{equation}
\label{equ:3-1-3}
\begin{aligned}
\log \sigma ((v')_{c_{t,i}}^{T} h_{t,i} ) + \sum_{x=1}^{r} \mathbb{E}_{c_x \sim p(x)}\left[ \log \sigma (-v')_{c_{t,i}}^{T} h_{t,i}  \right]
\end{aligned}
\end{equation}

In the equation, $c_{t,i}$ represents the medication, and $h_{t,i}$ is the context of $c_{t,i}$ in vector form. Negative samples are denoted as $c_x$. When generating negative samples, we add additional constraint for non-central words. 
For example, let $S_{drug}$ be the prescriptions of all patients. The negative samples are chosen from the set $S_{drug}-E_{y_i}$, based on the corresponding term frequencies. This is different from ``word2vec'' model, in which any non-central word can be a negative sample. 

\subsection{Anomaly Detection Based on Ranking Accumulation}
\label{subsec:4}

\begin{figure}[htbp]
\centerline{\includegraphics[width=0.5\textwidth]{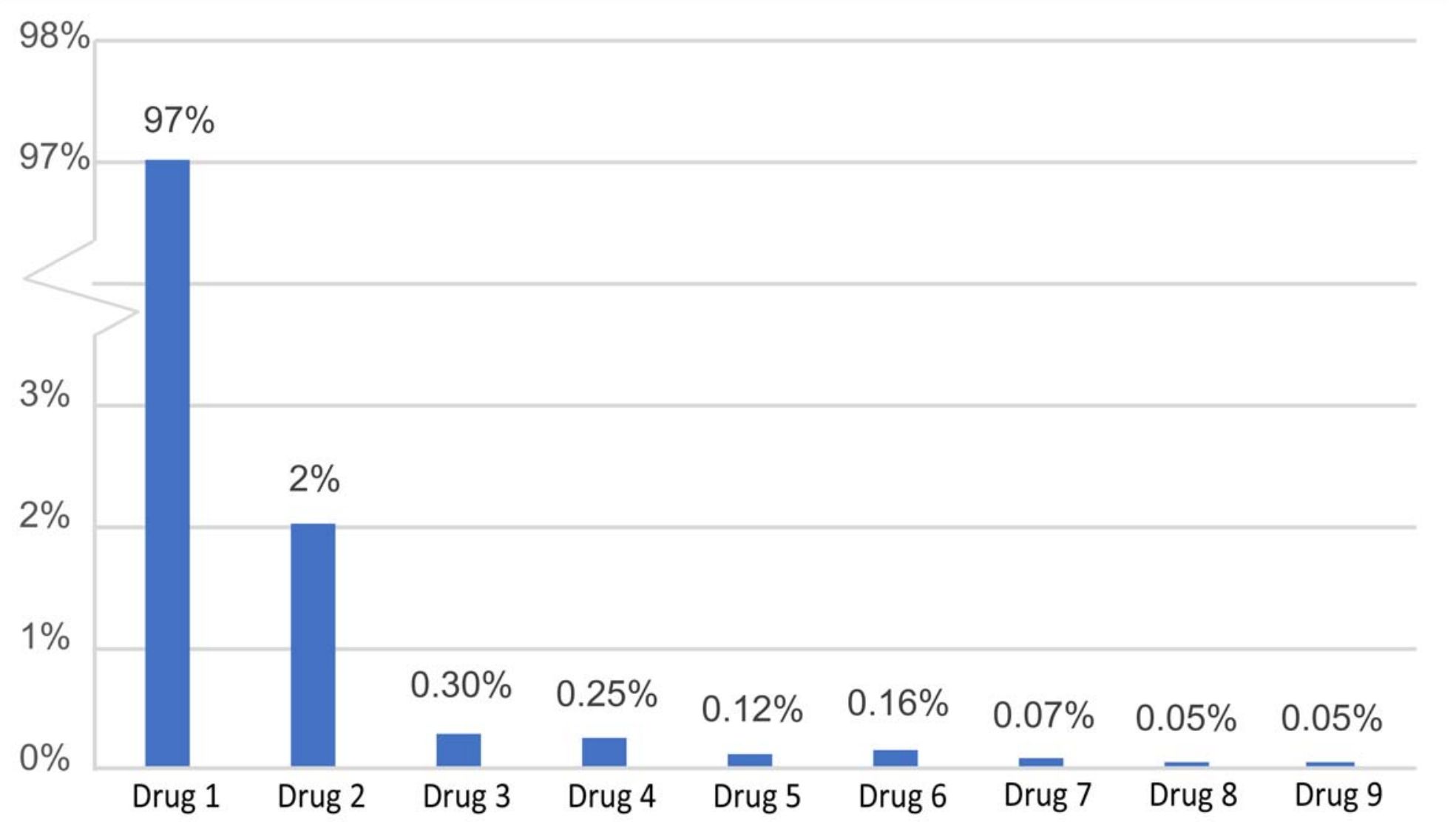}}
\caption{Examples of ranking scores.}
\label{fig2}
\end{figure}

With the model properly trained, it is able to predict the central word with high accuracy. However, solely relying on predicting the central word to distinguish prescription error would lead to ambiguous results. Specifically, the CBOW predicts the central word in a relatively high probability once training is complete. Fig. \ref{fig2} is an example of such results, in which Drug 1 to 3 are normal drugs and Drug 4 is the anomaly. The model outputs Drug 1 as the central word with almost 100\% probability, leaving the rest of the candidates close to 0. Note that the medication that is wrongfully given to a patient is not necessary to be the central word. Instead, it is the least possible medication that occurs in patients' diagnosis. Therefore, the task to detect the anomaly equals to find the least possible word as the central word. In this case, we can only rule out Drug 1 as the anomaly, and the rest of the candidates are still indistinguishable. 

To tackle the problem, we applied a ranking strategy, called Ranking Accumulation (RA), in the process of anomaly detection. 
For a specific diagnosis, it is intuitive that the smaller the probability that a drug is predicted to be a normal drug (central word), the higher the probability that the drug will be the anomaly. 
For each training sample, we can generate the probability of different medications in the output layer. Let the numbers of surrounding words and diagnosis be $m$ and $n$, respectively. We use the following steps to help exaggerate the difference of probability. 
\begin{enumerate}
\item Selection of central words: select $k$ medications from the prescriptions $E_{y_i}$ as $E^k$ and $s=n-k$ negative candidate drugs as $E_{neg}^s$;
\item Selection of surrounding words: select m different samples from all diagnoses $D_{y_i}$ as $D^m$;
\item Ranking the medications: use $D^m$, $E^k$ and $E_{neg}^s$ as inputs to obtain the probabilities of each candidate medication as the central word. Use the output value as the ranking value for each of the candidates;
\item Accumulating the ranking scores: if patients have $M$ diagnoses, there will be $C_M^m$ ranking scores in total. By summing the scores of each medications, the final score of the drug, denoted as $SumRank_i$. The smaller value of $SumRank_i$ indicates a higher probability of anomaly.
\end{enumerate}

In terms of detecting medication anomaly, $k$ central words and $m-k$ negative samples were selected from the prescriptions of one patient, while only one central word and $m-1$ negative samples are to be chosen for conventional CBOW related tasks. In this way, the model can focus on medication used by patients, rather than influenced by the negative samples. 

\section{Experiments}
\label{sec:exp}

\subsection{Datasets and Experimental Setup}

All experiments are conducted on a X64 Unix Sever with a 3.6 GHz i7 CPU and 8GB RAM. 
Two datasets, namely SG1213 (cancer patients) and XS1213 (heart failure patients), are used in the experiments, which are collected by Shuguang Hospital Affiliated to Shanghai University of Traditional Chinese Medicine and formulated as a structural triple, denoted as $(patient, diagnosis, drugs)$. Other details of the datasets are shown as Table \ref{Tab:Datasets}. 

\begin{table}[htbp]
\caption{Datasets Specification}
\begin{center}
\newcommand{\tabincell}[2]{\begin{tabular}{@{}#1@{}}#2\end{tabular}}
\begin{tabular}{c|c|c|c|c}
\hline
\textbf{Dataset}& \textbf{No. of Records} & \textbf{Patients} & \textbf{Diagnosis} & \textbf{Drugs} \\
\hline
SG1213 & 970 & 203 & 440 & 578 \\
\hline
XS1213 & 7772 & 1001 & 643 & 1053 \\
\hline
\end{tabular}
\label{Tab:Datasets}
\end{center}
\end{table}

Characteristics, such as incompleteness, redundancy, and diversity, are shared in both datasets, resulting in four major obstacles in data preprocessing:
\begin{enumerate}
\item The symptoms in  Traditional Chinese Medicine (TCM)-related diagnoses and corresponding indications for Chinese patent medicines are relatively macroscopic and vague, leading to the relations between the two are not clear. For instance, EHR contains diagnoses like ``气血亏虚 (qi-blood deficiency)'' and Chinese patent medicines like ``冬凌草片 (rabdosia rubescens tablets)'';
\item Confusions exits in diagnoses, especially the naming of specific disease or symptoms. Since the diagnosis name is typed manually by the doctors, some descriptive sentences are inserted, such as ``右肺下叶腺癌T2N2M1b－IV期, 右肺门 (right lower lobe adenocarcinoma stage T2N2M1b-IV, right hilum of lung)''. The same diagnosis has multiple names, such as ``高血压病2级极高危 (extremely high risk of hypertension grade 2)'', ``高血压2级高危 (hypertension grade 2, high risk)'', and ``高血压病2级中危 (grade 2 medium-risk hypertension)'', all of which describe the condition ``高血压2级高危 (hypertension of high risk grade 2)'';
\item Coding system are chaotic. Both ICD10 and TCM codings exist in the data, resulting multiple codes for a single diagnoses. For example, $E11.900, E11.901, E14.901, E14.900$, and $E14.900$ are coded for the same disease $type 2 diabetes mellitus$. Meanwhile, one specific code may have diverse names or descriptions. For instance, ``支气管扩张伴感染 (bronchiectasis with infection)'' and ``支气管扩张 (bronchiectasis)'' both correspond to the code ``J47.01''. 
\item Rare drugs increase the difficulties of detecting the anomaly. Many medications in XS1213 and SG1213 appear in a relatively lower frequency ($\le 3$), leading conventional detection approaches to label them as anomalies. 
\end{enumerate}

In terms of the task described in this paper, we pre-processed the EHR data by ignoring the hospitalization time, and integrated diagnoses and prescriptions for each patient into sets $D_{y_i}$ and $E_{y_i}$, respectively. The former included all the diseases while the latter contained all the medications in the hospital records of the patients.

\subsubsection{Comparative Methods}
%
For distance-based methods, LOF was used to cluster the patients based on their diagnoses and prescriptions. The larger the LOF value is, the more likely the corresponding medication is the anomaly. 

For Naive Bayes (NB), we use two different settings. 
One is conventional NB, which assume that the diseases are independent. In this case, all the diagnoses and prescriptions are used to calculate the probabilities. 
However, considering that some diseases co-occur, we only use the diagnoses and prescriptions that frequently appear in the data. We call this setting NB+{$(i)$}, where $i=1,\cdots,5$, to distinguish with the conventional NB. 

For TransE, 
we define three relations as ``has\_disease'', ``has\_medicine'' and ``Corr'', which are used to describe the relations between disease, medication and patients. We use the diagnosis as the head entity (h), the medication as the tail (t), and the relation (r) has $h+r = t$. Distance between diagnose and prescriptions are calculated using the minimum value of all the distances between head and tail entities. Similar settings are also used for other Trans-series methods.


\subsubsection{Parameter Setting}
The parameters of the proposed method are shown in Table \ref{Tab:PrarameterSet}. We evaluated different number of surrounding words ranging from 1 to 10, and noted that, after $k > 3$, the performance remained still.
\begin{table}[htbp]
\caption{parameter settings}
\begin{center}
\newcommand{\tabincell}[2]{\begin{tabular}{@{}#1@{}}#2\end{tabular}}
\begin{tabular}{c|c}
\hline
\textbf{Parameter}& \textbf{Value}   \\
\hline
\tabincell{c}{No. of \\ Surrounding Words} & 2 \\
\hline
\tabincell{c}{No. of Negative Samples} & 5  \\
\hline
\tabincell{c}{No. of Central Words} & 1  \\
\hline
\tabincell{c}{No. of Diseases} & 64 \\
\hline
\end{tabular}
\label{Tab:PrarameterSet}
\end{center}
\end{table}
%
%
\subsubsection{Naming Formalization}
In this paper, we deleted the descriptive words related to stages and 
transformed the data into the form, saying ``左乳腺癌术后 (after left breast cancer surgery)'' and ``右肺下叶腺癌 (right lower lobe adenocarcinoma of lung)''. 
Similarly, hypertension is often described with a classification and risk degree, such as ``grade II'', ``medium risk'' and ``high risk'', which we unified into ``hypertension''. 
Other unifications are listed in Table \ref{Table_1}.

\begin{table}[htbp]
\caption{Unified Name for Patients' Diagnoses}
\begin{center}
\newcommand{\tabincell}[2]{\begin{tabular}{@{}#1@{}}#2\end{tabular}}
\begin{tabular}{c|c|c}
\hline
& \textbf{Unified Name} & \textbf{Original Name} \\
\hline
1 & \tabincell{c}{冠心病 \\ (
coronary heart disease)} & \tabincell{c}{ 冠状动脉粥样硬化性心脏病 \\ (Coronary atherosclerotic\\ heart disease)} \\
\hline
2 & \tabincell{c}{高血压 \\( Hypertension)} & \tabincell{c}{高血压3期（极高危）\\(Hypertension Phase 3,\\ Extremely High Risk),\\ 高血压病2级\\ (Hypertension Class 2) }\\
\hline
3 & \tabincell{c}{慢性心功能不全 \\ (Chronic cardiac\\ insufficiency)} & \tabincell{c}{慢性心功能不全 \\ 心功能III级(NYHA分级)\\(Class III of Chronic\\ Cardiac Insufficiency (NYHA))} \\
\hline
\end{tabular}
\label{Table_1}
\end{center}
\end{table}

\subsubsection{Medication Screening}
Some of the medications, eg. ``氯化钠针 (sodium chloride needles)'', appeared in almost every prescription given to patients, which we believe is most unlikely to be the anomaly. Therefore, we remove these records in the preprocessing of the dataset. 
Meanwhile, some medications (eg. ``脉络宁针(mailuoning needle)'' and ``脉络宁注射液(mailuoning injection)'') are given in diverse forms which should be unified (eg. ``脉络宁针(mailuoning needle'') as well. 


In order to evaluate the performance, we constructed an artificial standard answer set. Firstly, we obtained the corresponding indications for all drugs from the pharmacopeia website, before comparing the indications with the diagnoses of the patients. Finally, the indications for each drug that did not match the patients' diagnoses were regarded as abnormal drugs and were put into the standard answer set. 
Experienced doctors are included to help ensure the accuracy of the set.

\subsubsection{Evaluation Criteria}
Note that, for all the comparing methods, the outputs are lists of probabilities indicating the corresponding medications being the anomalies. 
To evaluate the performances, we choose $Top N$ accuracy, denoted as Equation (\ref{equ:topN}), which represents the percentage of anomalies exist in the first $N$ drugs. 
The letter $t$ in the equation is the number of anomalies detected by the methods, and $TN$ is the number of all anomalies in the prescriptions. 
By setting the values of $N$ ranging from $1$ to $5$, we can evaluate the accuracy of the methods in different granularities. 

\begin{equation}
\label{equ:topN}
\begin{aligned}
TopN = \frac{t}{TN}
\end{aligned}
\end{equation}

\subsection{Experimental Results}

The $Top~N$ accuracies on  SG1213 and XS1213 are presented in Table \ref{tab:topN_sg} and \ref{tab:topN_xs}, respectively. 
The proposed method, denoted as ``CBOWRA'', possesses the highest accuracy (over $0.84$) in all settings, 
Between the performances on two datasets, all methods on XS1213 performs slightly poor than on SG1213. We think the reason are two folds: The first is that medication are more diverse on XS1213 than on SG1213; The second is that the drugs with appearance less than $3$ occupy a higher portion in XS1213. 
The two reasons lead to an insufficient training of the models, thereby causes the performance to be less satisfying in XS1213.

\begin{table}[htbp]
\caption{Top N accuracies on SG1213}
\begin{center}
\newcommand{\tabincell}[2]{\begin{tabular}{@{}#1@{}}#2\end{tabular}}
\begin{tabular}{c|c|c|c|c|c}
\hline
\textbf{Method} & \textbf{Top 1} & \textbf{Top 2} & \textbf{Top 3} & \textbf{Top 4} & \textbf{Top 5} \\
\hline
LOF & 0.740 & 0.759 & 0.763 & 0.767 & 0.773 \\
\hline
NB & 0.410 & 0.405 & 0.456 & 0.498 & 0.528 \\
\hline
NB+(1) & 0.750 & 0.758 & 0.766 & 0.784 & 0.789 \\
\hline
NB+(2) & 0.755 & 0.775 & 0.792 & 0.796 & 0.793 \\
\hline
NB+(3) & 0.770 & 0.773 & 0.767 & 0.768 & 0.774 \\
\hline
NB+(4) & 0.690 & 0.695 & 0.688 & 0.696 & 0.699 \\
\hline
NB+(5) & 0.635 & 0.643 & 0.637 & 0.640 & 0.642 \\
\hline
TransE & 0.710 & 0.708 & 0.696 & 0.682 & 0.673 \\
\hline
TransH & 0.655 & 0.685 & 0.688 & 0.684 & 0.682 \\
\hline
TransD & 0.664 & 0.674 & 0.680 & 0.677 & 0.673 \\
\hline
TransR & 0.663 & 0.659 & 0.650 & 0.645 & 0.638 \\
\hline
CBOWRA & \textbf{0.879} & \textbf{0.867} & \textbf{0.858} & \textbf{0.851} & \textbf{0.844} \\
\hline
\end{tabular}
\label{tab:topN_sg}
\end{center}
\end{table}

In the results of NB, we also found that the model with $K = 2$ achieves the best performance, which gave us insights to set a proper length of surrounding words. In the experiments, we set the length of surrounding words to $2$ when training CBOW related methods.

\begin{table}[htbp]
\caption{Top N accuracies on XS1213}
\begin{center}
\newcommand{\tabincell}[2]{\begin{tabular}{@{}#1@{}}#2\end{tabular}}
\begin{tabular}{c|c|c|c|c|c}
\hline
\textbf{Method} & \textbf{Top 1} & \textbf{Top 2} & \textbf{Top 3} & \textbf{Top 4} & \textbf{Top 5} \\
\hline
LOF & 0.599 & 0.596 & 0.611 & 0.613 & 0.617 \\
\hline
NB & 0.434 & 0.543 & 0.544 & 0.560 & 0.561 \\
\hline
NB+(1) & 0.629 & 0.620 & 0.616 & 0.611 & 0.612 \\
\hline
NB+(2) & 0.607 & 0.615 & 0.610 & 0.610 & 0.608 \\
\hline
NB+(3) & 0.605 & 0.609 & 0.606 & 0.604 & 0.599 \\
\hline
NB+(4) & 0.616 & 0.611 & 0.599 & 0.596 & 0.593 \\
\hline
NB+(5) & 0.615 & 0.595 & 0.594 & 0.592 & 0.593 \\
\hline
TransE & 0.620 & 0.625 & 0.621 & 0.616 & 0.611 \\
\hline
TransH & 0.619 & 0.621 & 0.622 & 0.617 & 0.613\\
\hline
TransD & 0.575 & 0.591 & 0.592 & 0.595 & 0.598 \\
\hline
TransR & 0.638 & 0.623 & 0.620 & 0.618 & 0.613 \\
\hline
CBOWRA & \textbf{0.639} & \textbf{0.631} & \textbf{0.627} & \textbf{0.624} & \textbf{0.622} \\
\hline
\end{tabular}
\label{tab:topN_xs}
\end{center}
\end{table}

When using CBOW to cluster similar patients, the patient vectors are coded in multi-hot forms, with dimensions of $326$ (SG1213) and $643$ (XS1213), respectively. However, there are only $15$ diagnoses per patient on average, which makes the sparseness of patient vectors reach $95\%$ and $97\%$, respectively. Sparse vectors seriously affect the performance of clustering and the accuracy of the detection will be further reduced.

Trans-series is suitable for dealing with data with various kinds of relationships. However, there are only three kinds of entities in both of the datasets, namely ``patient'', ``medication'', and ``diagnosis'', and two kinds of relationships between entities, namely the relationship between patients and drugs and the one between patients and diagnoses. In addition, the relationships between entities are one-to-many, many-to-one, and many-to-many, which make the model ineffective. For example, with training data $\left \langle Patient A, Has\_drug, Drug 1 \right \rangle$, $	\left \langle Patient A, Has\_drug, Drug 2 \right \rangle $, both of them have the same head entity (h) and relation (r). TransE tends to treat the two tail entities, namely Drug1 and Drug2, as the same entity, which reduces the accuracy of the model and confuses correct and abnormal drugs.

\section{Discussion}

In the experiments, the proposed method has achieved competitive performance in all settings. 
However, the datasets only contain two medical concepts, namely diagnosis and medication, and the performances on specific diagnoses are still less satisfying. 
By analyzing the data and the aforementioned results, we summarize the reasons as follows.
\begin{enumerate}
\item 
Considerable amount of the prescriptions are injections, which are commonly used for surgical hemostasis, analgesia, suppression of side effects of radiotherapy and chemotherapy, and adjuvant therapy. The data only appear in the EHR when doctors are operating on the patients, and record in a non-real time manner. Meanwhile, doctors usually do not record the symptoms of patients during surgery, nor do they record whether they have undergone radiotherapy or chemotherapy. Therefore, these medications are classified to be the anomaly, despite the fact they are indeed appropriate medications and appear in almost every patient's prescription. 
\item 
In the datasets, some anti-side-effect medications are considered to be anomalies because they do not have direct connection with the disease. For example, patients treated with chemotherapy or radiotherapy have constipation as a side effect. In the dataset, suppositories glycerol, which is used to treat the symptom, is labeled as an anomaly. However, with the proposed model trained to discover semantic consistency over the diagnosis and prescription, these medications are identified as normal drugs. Therefore, some of the mismatches confirms the effectiveness of the proposed method.
\item 
Some patients use rare drugs. By ``rare'', it means there are few people using this kind of drug and they should not be considered as an anomaly as other outlier detection methods do. Due to insufficient data volume, the proposed model fails to distinguish such medication from the true anomaly.
\end{enumerate}

The future work lies in two aspects. The first is to improve the construction of standard answers. In the paper, we used the relevant knowledge on encyclopedia websites and a string similarity matching method in the construction. However, it does not rule out factors such as low accuracy, incomplete information, and poor timeliness of medical knowledge, all of which can interfere with the accuracy of standard answers. The other extension of the work is to 
use information, such as patient examinations and operations, to enrich the context of prescriptions, which we believe will further improve the medication anomaly detection.

\section{Conclusion}
In this manuscript, we present a simple, yet effective method, i.e. CBOWRA, to detect possible medication anomaly in EHR systems, which adopts representation learning to reveal the semantic consistency between diagnoses and prescriptions in patients' files. Specifically, we combine CBOW and a ranking accumulation strategy to construct the model, and evaluate the performances with other methods ranging from conventional outlier detection to recent representation learning approaches. The experimental results of the proposed method outperforms the others by over 3.91\% and 0.68\% in two real hospital data, respectively. In addition, we also discussed the possible improvements and future directions for the proposed method.

\section*{Acknowledgment}
Firstly, the author would like to appreciate any suggestions and comments from the anonymous reviews and editor. 
The paper  is funded by the National Natural Science Foundation of China (Grant No. 61772201), the National Key R\&D Program of China for ``Precision Medical Research" (Grant No. 2018YFC0910500), and the Special Fund Project for ``Shanghai Informatization Development in Big Data" (Grant No. 201901043).


\end{CJK}
\bibliographystyle{IEEEtran}
\bibliography{IEEEabrv,mybib}

\begin{thebibliography}{10}
\providecommand{\url}[1]{#1}
\csname url@samestyle\endcsname
\providecommand{\newblock}{\relax}
\providecommand{\bibinfo}[2]{#2}
\providecommand{\BIBentrySTDinterwordspacing}{\spaceskip=0pt\relax}
\providecommand{\BIBentryALTinterwordstretchfactor}{4}
\providecommand{\BIBentryALTinterwordspacing}{\spaceskip=\fontdimen2\font plus
\BIBentryALTinterwordstretchfactor\fontdimen3\font minus
  \fontdimen4\font\relax}
\providecommand{\BIBforeignlanguage}[2]{{%
\expandafter\ifx\csname l@#1\endcsname\relax
\typeout{** WARNING: IEEEtran.bst: No hyphenation pattern has been}%
\typeout{** loaded for the language `#1'. Using the pattern for}%
\typeout{** the default language instead.}%
\else
\language=\csname l@#1\endcsname
\fi
#2}}
\providecommand{\BIBdecl}{\relax}
\BIBdecl

\bibitem{yadav2018mining}
P.~Yadav, M.~Steinbach, V.~Kumar, and G.~Simon, ``Mining electronic health
  records ({EHRs}): a survey,'' \emph{ACM Computing Surveys (CSUR)}, vol.~50,
  no.~6, p.~85, 2018.

\bibitem{liu2018patienteg}
X.~Liu, J.~Jin, Q.~Wang, T.~Ruan, Y.~Zhou, D.~Gao, and Y.~Yin, ``Patienteg
  dataset: Bringing event graph model with temporal relations to electronic
  medical records,'' \emph{arXiv preprint arXiv:1812.09905}, 2018.

\bibitem{zwaan2015challenges}
L.~Zwaan and H.~Singh, ``The challenges in defining and measuring diagnostic
  error,'' \emph{Diagnosis}, vol.~2, no.~2, pp. 97--103, 2015.

\bibitem{C1997Medication}
C.~Galper, ``Medication errors,'' \emph{The Lancet}, vol. 349, no. 9056, pp.
  959--960, 1997.

\bibitem{schiff2009diagnostic}
G.~D. Schiff, O.~Hasan, S.~Kim, R.~Abrams, K.~Cosby, B.~L. Lambert, A.~S.
  Elstein, S.~Hasler, M.~L. Kabongo, N.~Krosnjar \emph{et~al.}, ``Diagnostic
  error in medicine: analysis of 583 physician-reported errors,''
  \emph{Archives of internal medicine}, vol. 169, no.~20, pp. 1881--1887, 2009.

\bibitem{aggarwal2015outlier}
C.~C. Aggarwal, ``Outlier analysis,'' in \emph{Data mining}.\hskip 1em plus
  0.5em minus 0.4em\relax Springer, 2015, pp. 237--263.

\bibitem{tran2016distance}
L.~Tran, L.~Fan, and C.~Shahabi, ``Distance-based outlier detection in data
  streams,'' \emph{Proceedings of the VLDB Endowment}, vol.~9, no.~12, pp.
  1089--1100, 2016.

\bibitem{li2015privacy}
L.~Li, L.~Huang, W.~Yang, X.~Yao, and A.~Liu, ``Privacy-preserving lof outlier
  detection,'' \emph{Knowledge and Information Systems}, vol.~42, no.~3, pp.
  579--597, 2015.

\bibitem{swersky2016evaluation}
L.~Swersky, H.~O. Marques, J.~Sander, R.~J. Campello, and A.~Zimek, ``On the
  evaluation of outlier detection and one-class classification methods,'' in
  \emph{IEEE International Conference on Data Science and Advanced Analytics
  (DSAA)}.\hskip 1em plus 0.5em minus 0.4em\relax IEEE, 2016, pp. 1--10.

\bibitem{campos2016evaluation}
G.~O. Campos, A.~Zimek, J.~Sander, R.~J. Campello, B.~Micenkov{\'a},
  E.~Schubert, I.~Assent, and M.~E. Houle, ``On the evaluation of unsupervised
  outlier detection: measures, datasets, and an empirical study,'' \emph{Data
  Mining and Knowledge Discovery}, vol.~30, no.~4, pp. 891--927, 2016.

\bibitem{domingues2018comparative}
R.~Domingues, M.~Filippone, P.~Michiardi, and J.~Zouaoui, ``A comparative
  evaluation of outlier detection algorithms: Experiments and analyses,''
  \emph{Pattern Recognition}, vol.~74, pp. 406--421, 2017.

\bibitem{Capelleveen2016Outlier}
G.~V. Capelleveen, M.~Poel, R.~M. Mueller, D.~Thornton, and J.~V.
  Hillegersberg, ``Outlier detection in healthcare fraud: A case study in the
  medicaid dental domain,'' \emph{International Journal of Accounting
  Information Systems}, vol.~21, pp. 18--31, 2016.

\bibitem{goedecke2016medication}
T.~Goedecke, K.~Ord, V.~Newbould, S.~Brosch, and P.~Arlett, ``Medication
  errors: new eu good practice guide on risk minimisation and error
  prevention,'' \emph{Drug safety}, vol.~39, no.~6, pp. 491--500, 2016.

\bibitem{schiff2017screening}
G.~D. Schiff, L.~A. Volk, M.~Volodarskaya, D.~H. Williams, L.~Walsh, S.~G.
  Myers, D.~W. Bates, and R.~Rozenblum, ``Screening for medication errors using
  an outlier detection system,'' \emph{Journal of the American Medical
  Informatics Association}, vol.~24, no.~2, pp. 281--287, 2017.

\bibitem{zhang2016probabilistic}
L.~Zhang, X.~Li, H.~Liu, J.~Mei, G.~Hu, J.~Zhao, Y.~Zou, B.~Xie, and G.~Xie,
  ``Probabilistic-mismatch anomaly detection: Do one's medications match with
  the diagnoses,'' in \emph{2016 IEEE 16th International Conference on Data
  Mining (ICDM)}.\hskip 1em plus 0.5em minus 0.4em\relax IEEE, 2016, pp.
  659--668.

\bibitem{taylor2018predicting}
R.~A. Taylor, C.~L. Moore, K.-H. Cheung, and C.~Brandt, ``Predicting urinary
  tract infections in the emergency department with machine learning,''
  \emph{PloS one}, vol.~13, no.~3, p. e0194085, 2018.

\bibitem{goh2018drug}
W.~P. Goh, X.~Tao, J.~Zhang, J.~Yong, W.~Zhang, and H.~Xie, ``Drug prescription
  support in dental clinics through drug corpus mining,'' \emph{International
  Journal of Data Science and Analytics}, vol.~6, no.~4, pp. 341--349, 2018.

\bibitem{cai2016natural}
T.~Cai, A.~A. Giannopoulos, S.~Yu, T.~Kelil, B.~Ripley, K.~K. Kumamaru, F.~J.
  Rybicki, and D.~Mitsouras, ``Natural language processing technologies in
  radiology research and clinical applications,'' \emph{Radiographics},
  vol.~36, no.~1, pp. 176--191, 2016.

\bibitem{pons2016natural}
E.~Pons, L.~M. Braun, M.~M. Hunink, and J.~A. Kors, ``Natural language
  processing in radiology: a systematic review,'' \emph{Radiology}, vol. 279,
  no.~2, pp. 329--343, 2016.

\bibitem{yang2018learning}
Y.~Yang, S.~Yuan, D.~Cer, S.-y. Kong, N.~Constant, P.~Pilar, H.~Ge, Y.-H. Sung,
  B.~Strope, and R.~Kurzweil, ``Learning semantic textual similarity from
  conversations,'' \emph{arXiv preprint arXiv:1804.07754}, 2018.

\bibitem{gao2017video}
L.~Gao, Z.~Guo, H.~Zhang, X.~Xu, and H.~T. Shen, ``Video captioning with
  attention-based lstm and semantic consistency,'' \emph{IEEE Transactions on
  Multimedia}, vol.~19, no.~9, pp. 2045--2055, 2017.

\bibitem{xie2018learning}
S.~Xie, Z.~Zheng, L.~Chen, and C.~Chen, ``Learning semantic representations for
  unsupervised domain adaptation,'' in \emph{International Conference on
  Machine Learning}, 2018, pp. 5419--5428.

\bibitem{qiu2018fast}
J.~Qiu, Q.~Wang, Y.~Zhou, T.~Ruan, and J.~Gao, ``Fast and accurate recognition
  of chinese clinical named entities with residual dilated convolutions,'' in
  \emph{2018 IEEE International Conference on Bioinformatics and Biomedicine
  (BIBM)}.\hskip 1em plus 0.5em minus 0.4em\relax IEEE, 2018, pp. 935--942.

\bibitem{mikolov2013efficient}
T.~Mikolov, K.~Chen, G.~Corrado, and J.~Dean, ``Efficient estimation of word
  representations in vector space,'' \emph{arXiv preprint arXiv:1301.3781},
  2013.

\bibitem{bordes2013translating}
A.~Bordes, N.~Usunier, A.~Garcia-Duran, J.~Weston, and O.~Yakhnenko,
  ``Translating embeddings for modeling multi-relational data,'' in
  \emph{Advances in neural information processing systems}, 2013, pp.
  2787--2795.

\bibitem{wang2014knowledge}
Z.~Wang, J.~Zhang, J.~Feng, and Z.~Chen, ``Knowledge graph embedding by
  translating on hyperplanes,'' in \emph{Twenty-Eighth AAAI conference on
  artificial intelligence}, 2014.

\bibitem{ji2015knowledge}
G.~Ji, S.~He, L.~Xu, K.~Liu, and J.~Zhao, ``Knowledge graph embedding via
  dynamic mapping matrix,'' in \emph{Proceedings of the 53rd Annual Meeting of
  the Association for Computational Linguistics and the 7th International Joint
  Conference on Natural Language Processing (Volume 1: Long Papers)}, 2015, pp.
  687--696.

\bibitem{lin2015learning}
Y.~Lin, Z.~Liu, M.~Sun, Y.~Liu, and X.~Zhu, ``Learning entity and relation
  embeddings for knowledge graph completion,'' in \emph{Twenty-ninth AAAI
  conference on artificial intelligence}, 2015.

\bibitem{mikolov2013distributed}
T.~Mikolov, I.~Sutskever, K.~Chen, G.~S. Corrado, and J.~Dean, ``Distributed
  representations of words and phrases and their compositionality,'' in
  \emph{Advances in neural information processing systems}, 2013, pp.
  3111--3119.

\end{thebibliography}

\end{document}